\documentclass[11pt]{article}
\usepackage{iclr2026_conference,times}
\usepackage[margin=1in]{geometry}
\usepackage{amsmath,amssymb}
\usepackage{booktabs}
\usepackage{multirow}
\usepackage{makecell}
\usepackage{tabularx}
\usepackage{graphicx}
\usepackage{subcaption}
\usepackage{xcolor}
\usepackage{hyperref}
\usepackage[nameinlink,capitalise]{cleveref}
\usepackage{natbib}
\usepackage{url}
\usepackage{fancyvrb}
\usepackage{fvextra}
\usepackage{multicol}
\usepackage{enumitem}
\usepackage{xurl}
\usepackage[normalem]{ulem}

\usepackage[most]{tcolorbox}
\usepackage{listings}

\newcommand{\op}[1]{\item \path{#1}}
\definecolor{crimson}{HTML}{DC143C}

\usepackage{amsmath}
\usepackage{amsfonts}
\usepackage{amssymb}
\usepackage{booktabs}
\usepackage{multirow}
\usepackage{makecell}
\usepackage{array}
\usepackage{graphicx}
\usepackage{booktabs}
\usepackage{multirow}
\usepackage{makecell}
\usepackage{threeparttable}
\usepackage{adjustbox}
\usepackage[table]{xcolor}
\usepackage{siunitx}

\definecolor{oraclegray}{RGB}{246,246,246}
\definecolor{macrogray}{gray}{0.95}

\iclrfinalcopy

\definecolor{mtblue}{HTML}{C2410C}

\newcommand{\makeMusaTitle}{
\begin{center}

\vspace*{-6.3em}

\begin{minipage}{\textwidth}
    \raggedright
    \includegraphics[height=0.65cm]{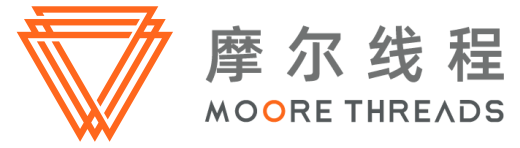}
\end{minipage}

\rule{\textwidth}{0.8pt}

{\Large\bfseries Beyond the Best Teacher: Expanding and Compressing\\[0.4em]
the Reasoning Solution Manifold}
\vspace{1.0em}

{\large\bfseries
Songshuo Lu\quad
Zhi Chen\quad
Yaohua Tang
}

\vspace{1.0em}

{\normalsize
Moore Threads AI\\[0.25em]
\texttt{tangyaohua28@gmail.com}
}

\vspace{0.6em}

\vspace{1.5em}

\end{center}
}

\title{
\includegraphics[height=1.2cm]{figures/logo.png}\\[1.0em]
Beyond the Best Teacher: Expanding and Compressing the Reasoning Solution Manifold}

\author{%
    Songshuo Lu \quad
    Zhi Chen \quad
    Yaohua Tang \\
    \\[0.2em]
    Moore Threads AI \\
    \texttt{tangyaohua28@gmail.com}
}
\date{}

\begin{document}
% \maketitle
\makeMusaTitle

\begin{abstract}
A single reinforcement-learning run can produce a strong reasoner yet an incomplete teacher: it often amplifies only a subset of the valid solution modes.
We argue that reinforcement learning (RL)-trained policies should therefore be viewed as local probes of a multi-basin reasoning solution manifold, rather than as globally reliable supervisors.
Based on this view, we propose an \emph{expand-then-compress} framework that couples teacher construction with multi-teacher policy distillation.
In the expansion stage, Residual Group Relative Policy Optimization (RGRPO) trains a sequence of teachers from a common initialization and redirects each later round toward examples not yet covered by the accumulated teacher union.
In the compression stage, reliability-gated Teacher-Union On-policy Distillation (TU-OPD) lets the student learn from its own response prefixes.
For each example, only reliable teachers contribute, and their sampled-token OPD losses are weighted by their per-example quality.
We further introduce Consensus-Residual Decomposition, which preserves a winner teacher's excess token preferences over its reliable peers, preventing specialist behavior from being suppressed during teacher aggregation.
Experiments on mathematical reasoning, code generation, and instruction following show that the resulting Qwen3-1.7B student consistently outperforms the strongest individual teacher across all three domains, yielding relative improvements of $2.0\%$, $8.3\%$, and $6.9\%$, respectively, while retaining single-model inference.
These results establish a simple but powerful principle: stronger students can be obtained not by selecting a single better teacher, but by deliberately constructing and compressing a complementary teacher union.
\end{abstract}


\section{Introduction}

\begin{figure*}[t] 
    \centering
    \includegraphics[width=\textwidth]{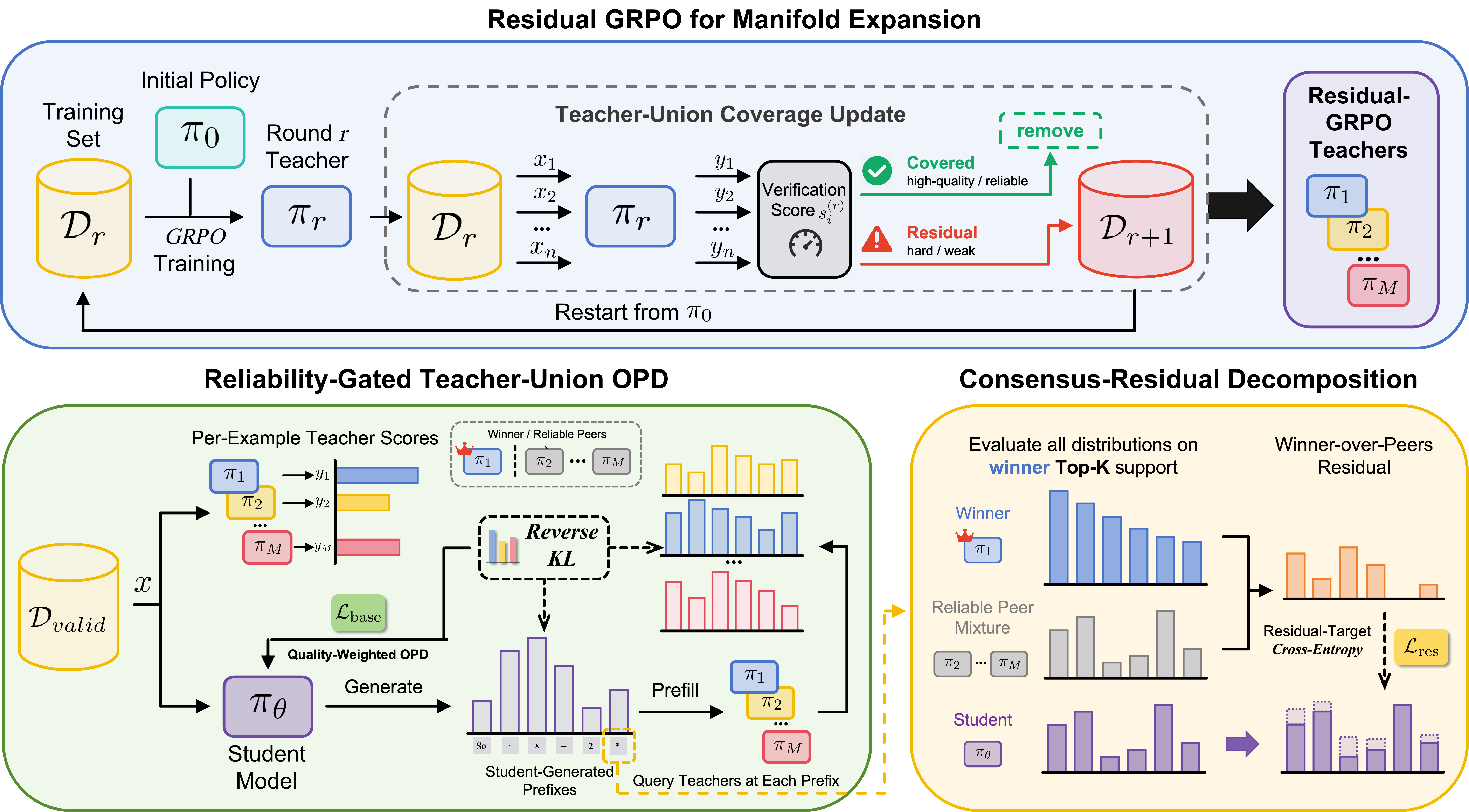}
    \caption{Overview of our expand-then-compress framework.
Residual GRPO training expands the reasoning solution manifold by training complementary teachers on subsets not yet covered by the teacher union.
Reliability-gated Teacher-Union OPD then compresses their reliable on-policy supervision into a single student through $\mathcal{L}_{\mathrm{base}}$.
Finally, Consensus-Residual Decomposition extracts the winner teacher's excess Top-$K$ token preference over its reliable peers and transfers it through $\mathcal{L}_{\mathrm{res}}$.}

    \label{fig:pipeline}
\end{figure*}

Reinforcement learning (RL) with verifiable rewards has become a powerful recipe for improving reasoning language models. Methods such as Group Relative Policy Optimization (GRPO) have enabled open models to make substantial progress on mathematical reasoning \cite{shao2024deepseekmath}, and large-scale RL has further shown that outcome rewards can elicit long-form reasoning and self-correction behaviors \cite{guo2025deepseek}. At the same time, recent studies on reasoning data and test-time scaling suggest that reasoning performance depends not only on model size or training compute, but also on how effectively training exposes and consolidates useful solution patterns \cite{muennighoff2025s1,ye2025limo,guha2025openthoughts}.

A key limitation of standard outcome-driven RL is that it treats policy improvement as if there were a single target behavior to discover. In complex reasoning tasks, however, correct solutions are rarely unique. The same problem can often be solved through different decompositions, intermediate representations, and reasoning strategies. We refer to this multi-basin structure as the reasoning solution manifold. A single RL run may amplify one discovered basin while leaving other valid basins under-explored. As a result, an RL-trained model can become a strong but incomplete teacher: reliable on the solution modes it has captured, yet brittle on regions of the training distribution that require alternative modes.

This observation motivates a different view of RL-trained teachers. Rather than treating a teacher as a globally reliable supervisor, we view it as a local probe of the solution manifold. Once a teacher has covered part of the training distribution, continuing to spend RL compute on the same covered region provides limited marginal value. Instead, later teachers should be directed toward the residual regions not yet covered by the current teacher union. The challenge is then twofold: how to expand coverage of the solution manifold with multiple partial teachers, and how to compress their complementary capabilities back into a single policy.

We propose \textbf{Expanding and Compressing the Reasoning Solution Manifold}, an expand-then-compress framework for reasoning policy distillation. In the expansion stage, we train a sequence of GRPO teachers from a common initialization. After each teacher is trained, we score its response quality on the full training set and maintain a coverage envelope over the teacher union. Examples already covered by the union are excluded from later GRPO rounds, while uncovered examples form the residual training subset for the next teacher. This produces teachers that are not merely sequential checkpoints, but specialists induced by different residual regions of the training distribution.

In the compression stage, we distill the resulting teacher union into a single student through online policy distillation \cite{agarwal2024policy,lu2025onpolicydistillation,li2026rethinking}. For each example, only teachers that reliably cover that example contribute to the base distillation objective. This reliability-gated teacher-union distillation lets the student learn from the appropriate local teachers rather than from a fixed global mixture. To avoid collapsing the compression process into teacher consensus, we further introduce a consensus-residual decomposition: the base objective transfers shared teacher behavior, while a residual objective preserves the winner teacher's excess preference relative to its reliable peers.

Our approach differs from prior multi-teacher distillation, which mainly focuses on aggregating or selecting supervision from an existing teacher set \cite{tian2025beyond,li2025learning,jin2026exploring,ma2026mopd}. Here, teacher construction and teacher compression are coupled. Residual GRPO (RGRPO) expands the empirical coverage of the reasoning solution manifold, and Teacher-Union On-policy Distillation (TU-OPD) compresses this expanded coverage into one policy without discarding non-consensus specialist components.

Our contributions are summarized as follows:
\begin{itemize}
    \item We introduce an \emph{expand-then-compress} framework for
    reasoning policy learning, viewing RL-trained policies as local probes
    of a multi-basin reasoning solution manifold rather than as globally
    complete teachers.

    \item We propose \emph{Residual GRPO} for manifold expansion.
    Starting from a common initialization, it constructs complementary
    specialists by redirecting each subsequent RL round toward examples
    not yet reliably covered by the accumulated teacher union.

    \item For manifold compression, we develop
    \emph{Reliability-Gated Teacher-Union OPD}, which selects and weights
    teachers according to their per-example competence, together with
    \emph{Consensus-Residual Decomposition}, which preserves
    winner-specific token preferences that conventional teacher
    aggregation may suppress.
\end{itemize}

Experiments across mathematical reasoning, code generation, and instruction following show that deliberately expanding teacher complementarity before compression yields a single student that consistently surpasses every individual teacher.

\section{Related Work}

\paragraph{Reasoning with verifiable rewards.}
Recent reasoning models increasingly rely on outcome-driven reinforcement learning with verifiable rewards. DeepSeekMath introduced GRPO for mathematical reasoning without an explicit value model~\citep{shao2024deepseekmath}, while DeepSeek-R1 demonstrated that large-scale RL can substantially improve complex reasoning behavior~\citep{guo2025deepseek}. Related work further studies how test-time scaling, data quality, and training recipes affect reasoning performance~\citep{muennighoff2025s1,ye2025limo,guha2025openthoughts}. However, these approaches mainly optimize a single policy over a fixed training distribution. Under sparse outcome rewards and limited exploration, one RL run may repeatedly reinforce discovered trajectories while leaving alternative valid strategies underexplored. Policies trained from different subsets may therefore develop distinct strengths even when initialized from the same base model, yet such variation is typically treated as optimization randomness or resolved by selecting only the strongest policy.

\paragraph{On-policy distillation.}
On-policy distillation (OPD) addresses the mismatch between offline teacher trajectories and the student's own state distribution by querying supervision on student-generated prefixes~\citep{agarwal2024policy,lu2025onpolicydistillation}. Recent studies extend OPD to reasoning, long-context modeling, self-distillation, context distillation, and adaptive token-level transfer~\citep{li2026rethinking,zhang2026opsdl,tan2026self,ye2026policy,zhao2026rosd,jin2026entropy,lee2026sead,xu2026tip,xie2026llm,zhang2026egad}. These works improve the alignment between teacher supervision and the states actually visited by the student, but generally assume a teacher or teacher pool constructed independently of the subsequent distillation process. In multi-teacher settings, teacher usefulness can also vary across examples, making global teacher ranking or uniform aggregation suboptimal.

\paragraph{Multi-teacher distillation.}
Multi-teacher distillation aims to integrate knowledge from multiple teachers into a single deployable student. In reasoning settings, prior work transfers reasoning capabilities from multiple teachers to smaller models~\citep{tian2025beyond}, uses committee-style peer review to refine teacher signals~\citep{li2025learning}, and explores knowledge purification and multi-teacher on-policy distillation~\citep{jin2026exploring,ma2026mopd}. These methods mainly focus on selecting, filtering, or aggregating supervision from an existing teacher set. However, teacher competence is often example-dependent, and conventional aggregation may introduce unreliable supervision or suppress localized preferences unique to individual specialists. Consequently, increasing the number of teachers does not necessarily preserve their complementary capabilities in the student.

Our framework connects teacher construction and student compression. Residual GRPO constructs complementary teachers by redirecting successive RL rounds toward examples not yet covered by the accumulated teacher set. Reliability-Gated Teacher-Union OPD then aggregates reliable on-policy supervision, while Consensus-Residual Decomposition preserves winner-specific preferences that would otherwise be weakened during teacher aggregation. The resulting objective is to distill the capability union of deliberately diversified teachers rather than merely approximate their average behavior.

\section{Methodology}

\subsection{Overview}

For a reasoning problem \(x\) and a candidate response \(y\), let \(R(x,y)\in\{0,1\}\) denote an outcome verifier that indicates whether \(y\) correctly solves \(x\). Complex reasoning tasks often admit multiple verifier-approved responses, spanning different reasoning strategies, decompositions, intermediate representations, and solution styles. We view these diverse solution modes as forming a multi-basin \emph{reasoning solution manifold}. Although outcome-driven reinforcement learning can amplify successful trajectories, limited exploration, sparse rewards, and decreasing policy entropy often cause a single GRPO run to cover only a local region of this manifold.

Motivated by this perspective, we introduce an \emph{expand-then-compress} framework. In the expansion stage, Residual GRPO trains a sequence of teachers from a common initialization and directs each successive teacher toward examples not yet reliably covered by the accumulated teacher union. In the compression stage, Reliability-Gated Teacher-Union OPD consolidates their complementary capabilities into a single student using supervision queried on student-generated prefixes. To prevent teacher aggregation from suppressing specialist behavior, Consensus-Residual Decomposition further transfers winner-specific token preferences that exceed those of the winner's reliable peers.

\subsection{Residual GRPO for Manifold Expansion}

Given a training set
\begin{equation}
\mathcal{D}=\{x_i\}_{i=1}^{N},
\end{equation}
we construct \(M\) teacher policies
\begin{equation}
\mathcal{T}=\{\pi_1,\ldots,\pi_M\}.
\end{equation}
The first teacher is trained on the full training set:
\begin{equation}
\mathcal{D}_1=\mathcal{D}.
\end{equation}
At round \(r\), teacher \(\pi_r\) is trained from the common initialization policy \(\pi_0\) on the current subset \(\mathcal{D}_r\):
\begin{equation}
\pi_r=\operatorname{GRPO}(\pi_0;\mathcal{D}_r),
\qquad r=1,\ldots,M.
\end{equation}
All teachers are initialized from the same \(\pi_0\), rather than being obtained by continuously fine-tuning the previous teacher. This design makes the differences among teachers primarily attributable to their training subsets, instead of accumulated drift along a single optimization trajectory.

After training \(\pi_r\), we evaluate its response quality on the full training set \(\mathcal{D}\). 
For each teacher and example, we estimate its verification score using
$J$ sampled responses:
\begin{equation}
s_i^{(r)}
=
\frac{1}{J}
\sum_{j=1}^{J}
R\!\left(x_i,y_{i,j}^{(r)}\right),
\qquad
y_{i,j}^{(r)}
\sim\pi_r(\cdot\mid x_i).
\end{equation}

A larger value indicates that the teacher solves the example more reliably and is more likely to cover one basin of the corresponding solution manifold. The concrete estimation of \(s_i^{(r)}\) is described in the experimental setup.

Using these quality scores, we define the coverage envelope of the first \(r\) teachers on example \(x_i\) as
\begin{equation}
u_i^{(r)}=\max_{1\le a\le r}s_i^{(a)}.
\end{equation}
If \(u_i^{(r)}>\tau_{\mathrm{drop}}\), then \(x_i\) is considered covered by the current teacher union. Otherwise, it remains in the residual region. The next training subset is therefore
\begin{equation}
\mathcal{D}_{r+1}
=
\left\{
x_i\in\mathcal{D}:
u_i^{(r)}\le \tau_{\mathrm{drop}}
\right\}.
\end{equation}

This process is repeated for at most \(M\) rounds. Each new teacher is thus allocated to examples not yet covered by the existing teacher union. As a result, the teachers are not merely sequential checkpoints, but specialists induced by different residual regions of the training distribution.

\subsection{Reliability-Gated Teacher-Union OPD}

After constructing the teacher set, we compress its capability union into a student policy $\pi_\theta$.
We restrict distillation to
\begin{equation}
\mathcal{D}_{\mathrm{valid}}
=
\{x_i\in\mathcal{D}:u_i^{(M)}\ge\tau_{\mathrm{base}}\},
\end{equation}
and define the reliable teacher set for each retained example as
\begin{equation}
\mathcal{Q}_i
=
\{r:s_i^{(r)}\ge\tau_{\mathrm{base}}\}.
\end{equation}
The corresponding quality-aware teacher weights are
\begin{equation}
\alpha_i^{(r)}
=
\frac{\exp(s_i^{(r)}/T)}
{\sum_{a\in\mathcal{Q}_i}
 \exp(s_i^{(a)}/T)},
\qquad r\in\mathcal{Q}_i,
\end{equation}
where $T>0$ is the teacher-weighting temperature.

The student samples an on-policy response
$y_i\sim\pi_\theta(\cdot\mid x_i)$, with prefix
$h_{i,t}=(x_i,y_{i,<t})$.
Instead of computing full-vocabulary teacher distributions, the base
objective evaluates only the token $y_{i,t}$ sampled by the student:
\begin{equation}
\label{eq:sampled_token_base}
\mathcal{L}_{\mathrm{base}}
=
\mathbb{E}_{\substack{
x_i\sim\mathcal{D}_{\mathrm{valid}}\\
y_i\sim\pi_\theta(\cdot\mid x_i)
}}
\left[
\sum_t
\sum_{r\in\mathcal{Q}_i}
\alpha_i^{(r)}
\log
\frac{
\pi_\theta(y_{i,t}\mid h_{i,t})
}{
\pi_r(y_{i,t}\mid h_{i,t})
}
\right].
\end{equation}
Since $y_{i,t}$ is sampled from $\pi_\theta(\cdot\mid h_{i,t})$, this is a
Monte Carlo estimator of the quality-weighted reverse KL
$\sum_{r\in\mathcal{Q}_i}\alpha_i^{(r)}
D_{\mathrm{KL}}(\pi_\theta\Vert\pi_r)$.
It requires only the student and teacher log-probabilities of the sampled token.

\subsection{Consensus-Residual Decomposition}

The base OPD objective transfers behaviors shared by reliable teachers, but it can still suffer from consensus collapse during compression. When the advantage of a specialist teacher is concentrated in a small number of local decisions, averaging teacher distributions may suppress precisely those non-consensus preferences.

To address this, we operationally separate multi-teacher transfer into an aggregated component and a specialist residual component.
The base objective aggregates sampled-token OPD signals from reliable teachers, while the residual objective selectively transfers token preferences for which the strongest teacher exhibits a substantial advantage over its peers.

The base loss \(\mathcal{L}_{\mathrm{base}}\) aggregates the sampled-token OPD signals of reliable teachers. The residual objective captures the excess policy component of a winner teacher that cannot be explained by its peers.

For each example \(x_i\), we select the highest-quality reliable teacher:
\begin{equation}
r_i^\star=\arg\max_{r\in\mathcal{Q}_i}s_i^{(r)}.
\end{equation}
Residual supervision is enabled only when the winner has at least one reliable peer and its quality score exceeds the strongest peer by a clear margin:
\begin{equation}
z_i=
\mathbf{1}
\left[
|\mathcal{Q}_i|\ge 2
\ \land\
s_i^{(r_i^\star)}
-
\max_{a\in\mathcal{Q}_i\setminus\{r_i^\star\}} s_i^{(a)}
\ge \tau_{\mathrm{gap}}
\right].
\end{equation}
If no reliable peer exists, we set \(z_i=0\) and skip the residual objective.

Let $\mathcal{V}$ denote the model vocabulary.
For example $x_i$ and its student-generated prefix
$h_{i,t}=(x_i,y_{i,<t})$, the next-token probability assigned by
teacher $\pi_r$ to token $v\in\mathcal{V}$ is
\begin{equation}
p_{i,t}^{(r)}(v)
=
\pi_r(v\mid h_{i,t}),
\qquad v\in\mathcal{V}.
\end{equation}

We define the peer mixture for the winner teacher as
\begin{equation}
p_{i,t}^{\mathrm{peer}}(v)
=
\sum_{a\in\mathcal{Q}_i\setminus\{r_i^\star\}}
\bar{\alpha}_i^{(a)}p_{i,t}^{(a)}(v),
\end{equation}
where the peer weights are
\begin{equation}
\bar{\alpha}_i^{(a)}
=
\frac{\exp(s_i^{(a)}/T_{\mathrm{p}})}
{\sum_{b\in\mathcal{Q}_i\setminus\{r_i^\star\}}\exp(s_i^{(b)}/T_{\mathrm{p}})}.
\end{equation}
where $T_{\mathrm{p}}>0$ controls the concentration of the peer weights.

To focus residual transfer on the winner teacher's most probable local
decisions, we restrict the comparison to its Top-$K$ token support,
\begin{equation}
\mathcal{V}_{i,t}^{K}
=
\operatorname{TopK}_{K}
\left(
p_{i,t}^{(r_i^\star)}
\right)
\subseteq \mathcal{V},
\end{equation}
where $\operatorname{TopK}_{K}$ returns the set of $K$ tokens with the
largest probabilities under the winner teacher.

For each token $v\in\mathcal{V}_{i,t}^{K}$, we define its
winner-over-peer excess score as
\begin{equation}
\delta_{i,t}(v)
=
p_{i,t}^{(r_i^\star)}(v)
\left[
\log
\frac{
p_{i,t}^{(r_i^\star)}(v)+\epsilon
}{
p_{i,t}^{\mathrm{peer}}(v)+\epsilon
}
-\gamma
\right]_+,
% \qquad
% v\in\mathcal{V}_{i,t}^{K},
\end{equation}
where $[a]_+=\max(a,0)$, $\gamma\ge 0$ is a log-probability-ratio
margin, and $\epsilon>0$ is used for numerical stability.
This term becomes large only when the winner assigns substantial probability to token \(v\) and prefers it significantly more than the peer mixture. It therefore extracts a winner-specific residual component, rather than distilling the winner's full distribution.

We normalize the residual weights to obtain the residual target:
\begin{equation}
q_{i,t}(v)
=
\frac{\delta_{i,t}(v)}
{\sum_{u\in\mathcal{V}_{i,t}}\delta_{i,t}(u)+\epsilon},
\qquad v\in\mathcal{V}_{i,t}^{K}.
\end{equation}
The residual objective is
\begin{equation}
\mathcal{L}_{\mathrm{res}}
=
-
\mathbb{E}_{\substack{
x_i\sim\mathcal{D}_{\mathrm{valid}}\\
y_i\sim\pi_\theta(\cdot\mid x_i)
}}
\left[
z_i
\sum_{t=1}^{T_i}
\sum_{v\in\mathcal{V}_{i,t}^{K}}
q_{i,t}(v)
\log\pi_\theta(v\mid h_{i,t})
\right],
\end{equation}

The final student objective is
\begin{equation}
\mathcal{L}_{\mathrm{total}}
=
\mathcal{L}_{\mathrm{base}}
+
\lambda \mathcal{L}_{\mathrm{res}},
\end{equation}
where \(\lambda\) controls the strength of residual supervision.

\section{Experiments and Evaluation}
\label{sec:experiments}

\subsection{Experimental Setup}

\paragraph{Models and training data.}
We conduct our main experiments with Qwen3-1.7B and further evaluate
scalability with Qwen3-4B~\citep{yang2025qwen3}.
For each model scale, the initialization policy, Residual-GRPO teachers, and distilled student share the same model architecture.
The mathematical reasoning and code-generation training pools are constructed from Skywork-OR1-RL-Data~\citep{he2025skywork, skywork-or1-2025}, while the instruction-following experiments use the verifiable training data released with IFBench~\citep{pyatkin2025generalizing}.
For each task domain, we train four Residual-GRPO teachers from a common
initialization and subsequently compress their capability union into a
single student policy.

\paragraph{Benchmarks and evaluation.}

\begin{itemize}
    \item \textbf{Mathematical reasoning.}
    We evaluate on AIME 2024--2026~\citep{maa_aime}, HMMT February 2025, November 2025, and February 2026~\citep{hmmt_archive}, and AMC 2023~\citep{maa_amc}, and report Avg.@8 and Pass@8 for each benchmark.
    When AIME or HMMT is reported without an explicit year or month, it denotes the average over the three corresponding benchmarks; Math Mean is the unweighted average over all seven mathematical benchmarks.

    \item \textbf{Code generation.}
    We evaluate on the code-generation split of LiveCodeBench v6~\citep{jain2024livecodebench} and report Avg.@8 and Pass@8.

    \item \textbf{Instruction following.}
    We evaluate on the official IFBench test set~\citep{pyatkin2025generalizing} using its loose verifier and report Loose Avg.@8.
    Unless otherwise specified, all evaluation metrics are computed from eight sampled responses per problem.
\end{itemize}

\paragraph{Implementation details.}
We implement Residual GRPO and Teacher-Union OPD with VERL~\citep{sheng2024hybridflow}, and use vLLM for policy rollout and teacher inference~\citep{kwon2023efficient}.
Since all three task domains admit rule-based verification, we use binary outcome rewards \(R(x,y)\in\{0,1\}\) throughout training and data construction.
For both Residual GRPO and student distillation, we use a global batch size of \(32\), a learning rate of \(2\times10^{-6}\), one training epoch, and a maximum response length of \(32\)K tokens.
During Residual GRPO, we sample \(16\) responses per prompt for group-based policy optimization.
Separately, whenever teacher quality scores are estimated for coverage measurement or data filtering, we sample \(16\) responses from each teacher for every example and use their empirical success rate.
An example is removed from subsequent residual rounds once its teacher-union coverage exceeds \(\tau_{\mathrm{drop}}=0.5\).

For student distillation, the base Teacher-Union OPD objective is evaluated only on the token sampled by the student policy, whereas the residual objective operates on the winner teacher's Top-\(K\) support with \(K=8\). Teacher weights are normalized with a softmax temperature of \(0.25\).
Residual transfer is enabled when the winner teacher exceeds its strongest reliable peer by at least
\(\tau_{\mathrm{gap}}=0.2\), and its coefficient is fixed to
\(\lambda=0.05\).

\subsection{Main Results}

\begin{table*}[t]
\centering
\caption{
Main results on mathematical reasoning, code generation, and instruction following with Qwen3-1.7B.
}
\label{tab:main_results}

\small
\setlength{\tabcolsep}{2.5pt}
\renewcommand{\arraystretch}{1.12}

\sisetup{
    detect-weight=true,
    detect-family=true,
    table-number-alignment=center
}

\begin{adjustbox}{max width=\textwidth}
\begin{tabular}{
    l
    *{11}{S[table-format=2.2]}
}
\toprule

\multirow{3}{*}{\makecell[l]{Model\\(Qwen3-1.7B)}}
& \multicolumn{8}{c}{Math Reasoning (Avg.@8)}
& \multicolumn{2}{c}{Code Generation}
& \multicolumn{1}{c}{Instruction Following}
\\

\cmidrule(lr){2-9}
\cmidrule(lr){10-11}
\cmidrule(l){12-12}

& \multicolumn{3}{c}{AIME}
& \multicolumn{3}{c}{HMMT}
& \multicolumn{1}{c}{\multirow{2}{*}{AMC23}}
& \multicolumn{1}{c}{\multirow{2}{*}{\makecell{Math\\Mean}}}
& \multicolumn{2}{c}{LiveCodeBench v6}
& \multicolumn{1}{c}{\multirow{2}{*}{
    \makecell{IFBench\\Loose Avg.@8}
}}
\\

\cmidrule(lr){2-4}
\cmidrule(lr){5-7}
\cmidrule(lr){10-11}

&
\multicolumn{1}{c}{2024}
&
\multicolumn{1}{c}{2025}
&
\multicolumn{1}{c}{2026}
&
\multicolumn{1}{c}{Feb.\ 25}
&
\multicolumn{1}{c}{Nov.\ 25}
&
\multicolumn{1}{c}{Feb.\ 26}
&
&
&
\multicolumn{1}{c}{Avg.@8}
&
\multicolumn{1}{c}{Pass@8}
&
\\

\midrule

Initialization Policy ($\pi_0$)
& 1.25
& 1.67
& 1.67
& 2.50
& 1.67
& 1.52
& 27.50
& 5.40
& 4.70
& 15.26
& 17.67
\\

\midrule

Residual-GRPO Teacher 1
& 12.08
& 12.92
& \bfseries 12.92
& 6.67
& 8.75
& 6.82
& 48.13
& 15.47
& 19.23
& 31.75
& 26.67
\\

Residual-GRPO Teacher 2
& 13.75
& \bfseries 15.42
& 10.83
& 6.67
& 6.67
& 9.09
& 52.19
& 16.37
& 17.55
& 31.00
& 24.17
\\

Residual-GRPO Teacher 3
& \bfseries 15.42
& 12.08
& 7.50
& \bfseries 7.50
& \bfseries 10.00
& 8.33
& 51.25
& 16.01
& 18.50
& \bfseries 34.31
& 25.21
\\

Residual-GRPO Teacher 4
& 13.54
& 13.54
& 9.38
& 4.37
& 6.88
& \bfseries 9.47
& \bfseries 53.44
& 15.80
& 16.11
& 29.19
& 24.92
\\

\midrule

Avg.\ Residual-GRPO Teachers
& 13.70
& 13.49
& 10.16
& 6.30
& 8.08
& 8.43
& 51.25
& 15.91
& 17.85
& 31.56
& 25.24
\\

\rowcolor{gray!12}
Teacher Envelope$^\dagger$
& 15.42
& 15.42
& 12.92
& 7.50
& 10.00
& 9.47
& 53.44
& 17.74
& 19.23
& 34.31
& 26.67
\\

\midrule

\textbf{Ours}
& \bfseries 15.42
& 15.00
& 11.25
& 7.08
& 8.75
& 7.20
& 52.19
& \bfseries 16.70
& \bfseries 20.83
& 32.87
& \bfseries 28.50
\\

\bottomrule
\end{tabular}
\end{adjustbox}

\vspace{2pt}
\parbox{\textwidth}{
\small
$^\dagger$ Teacher Envelope reports the highest score among the four
Residual-GRPO teachers for each benchmark or metric.
Its Math Mean averages the seven benchmark-wise maxima.
Because its entries may come from different teachers, it is not a
deployable model.
}

\end{table*}

Table~\ref{tab:main_results} reports the initialization policy, the four
Residual-GRPO teachers, and our final distilled student.
We list all teachers individually to expose their complementary
benchmark-level strengths.
% \emph{Avg.\ Residual-GRPO Teachers} is the arithmetic mean of four
% independently evaluated models and does not involve inference-time
% ensembling.
We additionally report a \emph{Teacher Envelope}, which selects the
highest score among the four teachers separately for each benchmark or
evaluation metric.
Because its entries may come from different teachers, the envelope does
not correspond to a deployable model.

The individual teacher results reveal substantial specialization.
No single teacher dominates across the mathematical benchmarks:
the strongest results on AIME, HMMT, and AMC23 are distributed across
different Residual-GRPO rounds.
The same pattern appears on LiveCodeBench v6, where Teacher~1 achieves
the highest Avg.@8 among the teachers, whereas Teacher~3 obtains the
highest Pass@8.
Consequently, the Teacher Envelope reaches a Math Mean of $17.74$,
which is $1.37$ points, or $8.4\%$, above the strongest individual
teacher.
This gap provides direct benchmark-level evidence that Residual GRPO
constructs specialists with complementary capabilities rather than
multiple replicas of the same policy.

Our distilled student converts part of this complementarity into stronger
aggregate performance.
It achieves the highest deployable Math Mean of $16.70$, improving over
the strongest individual teacher by $2.0\%$ and over the teacher average
by $5.0\%$.
Although individual specialists remain stronger on several particular
benchmarks, the student recovers approximately $24\%$ of the gap between
the strongest mathematical teacher and the benchmark-wise Teacher
Envelope.
This shows that the student consolidates capabilities distributed across
teachers instead of merely reproducing the strongest specialist.

The consolidation benefit is also evident beyond mathematical reasoning.
On LiveCodeBench v6, our student reaches an Avg.@8 of $20.83$, a relative
improvement of $8.3\%$ over the strongest teacher on this metric.
Its Pass@8 of $32.87$ remains close to the best individual-teacher
result of $34.31$.
On IFBench, the student achieves a Loose Avg.@8 of $28.50$, exceeding the
strongest teacher by $6.9\%$ and the teacher average by $12.9\%$.

Overall, the student does not need to dominate every specialist on every
individual benchmark to demonstrate successful capability consolidation.
Instead, it achieves the strongest aggregate mathematical performance,
the highest LiveCodeBench Avg.@8, and the highest IFBench score among all
deployable models.
These results support our central claim: Teacher-Union OPD and
Consensus-Residual Decomposition can consolidate complementary
specialist capabilities into a single policy that is stronger on
task-level aggregate performance than any individual teacher.

\subsection{Ablation Study}

\begin{table*}[t]
\centering
\caption{
Capability-union complementarity of Residual-GRPO teachers and Parallel-GRPO teachers trained on four randomly partitioned, disjoint data subsets.
For Residual GRPO, columns report the cumulative teacher union after each round.
Union Avg.@8 selects the teacher with the highest per-example Avg.@8 before averaging over the evaluation set, while Union Pass@8 considers an example solved if any teacher produces at least one correct response.
}
\label{tab:residual_parallel_union}
% \small
% \setlength{\tabcolsep}{3.8pt}
% \renewcommand{\arraystretch}{1.12}

\setlength{\tabcolsep}{3.2pt}
\renewcommand{\arraystretch}{1.12}

\sisetup{
    detect-weight=true,
    detect-family=true,
    table-number-alignment=center
}

\begin{threeparttable}
\begin{tabular*}{\textwidth}{
    @{\extracolsep{\fill}}
    l
    *{10}{S[table-format=2.2]}
}
\toprule

\multirow{3}{*}{Benchmark}
& \multicolumn{8}{c}{Residual GRPO}
& \multicolumn{2}{c}{Parallel GRPO}
\\

\cmidrule(lr){2-9}
\cmidrule(l){10-11}

& \multicolumn{2}{c}{Round 1}
& \multicolumn{2}{c}{Rounds 1--2}
& \multicolumn{2}{c}{Rounds 1--3}
& \multicolumn{2}{c}{Rounds 1--4}
& \multicolumn{2}{c}{4 Teachers}
\\

\cmidrule(lr){2-3}
\cmidrule(lr){4-5}
\cmidrule(lr){6-7}
\cmidrule(lr){8-9}
\cmidrule(l){10-11}

& \multicolumn{1}{c}{Avg.@8}
& \multicolumn{1}{c}{Pass@8}
& \multicolumn{1}{c}{Avg.@8}
& \multicolumn{1}{c}{Pass@8}
& \multicolumn{1}{c}{Avg.@8}
& \multicolumn{1}{c}{Pass@8}
& \multicolumn{1}{c}{Avg.@8}
& \multicolumn{1}{c}{Pass@8}
& \multicolumn{1}{c}{Avg.@8}
& \multicolumn{1}{c}{Pass@8}
\\

\midrule

AIME 24--26
& 12.64
& 28.89
& \multicolumn{1}{c}{\underline{15.97}}
& \multicolumn{1}{c}{\underline{35.56}}
& 17.78
& 37.78
& \bfseries 18.61
& \bfseries 47.78
& 14.86
& 33.33
\\

HMMT 25--26
& 7.41
& 20.51
& 9.66
& 26.97
& 11.98
& \multicolumn{1}{c}{\underline{32.53}}
& \multicolumn{1}{c}{\underline{\bfseries 12.96}}
& {\bfseries 37.58}
& 12.64
& 32.22
\\

AMC23
& 48.13
& 87.50
& 55.94
& \multicolumn{1}{c}{\underline{92.50}}
& \multicolumn{1}{c}{\underline{59.38}}
& 92.50
& \bfseries 62.19
& \bfseries 92.50
& 58.13
& \bfseries 92.50
\\

\midrule

% \rowcolor{macrogray}
Macro Avg.$^\dagger$
& 22.73
& 45.63
& 27.19
& 51.68
& 29.71
& 54.27
& \bfseries 31.25
& \bfseries 59.29
& 28.54
& 52.68
\\

\bottomrule
\end{tabular*}

\vspace{2pt}
\parbox{\textwidth}{
\small
Underlined values indicate the earliest Residual-GRPO round whose
cumulative teacher union matches or surpasses the corresponding
four-teacher Parallel-GRPO union.

$^\dagger$ Macro Avg.\ is the unweighted average over the AIME, HMMT, and AMC23 task families.
}

\end{threeparttable}
\end{table*}

\paragraph{Teacher Complementarity: Does Residual GRPO Expand Coverage?}

We examine the expansion stage directly. Table~\ref{tab:residual_parallel_union} compares Residual GRPO with Parallel GRPO. In Parallel GRPO, the full training pool is randomly divided into four disjoint subsets, and one teacher is trained from the common initialization on each subset. This baseline creates diversity by data partitioning, but it does not adapt later teachers to the coverage of earlier ones.
In contrast, Residual GRPO constructs each subsequent subset from examples that remain insufficiently covered by the accumulated teacher union. 
This comparison isolates whether adaptive, coverage-aware allocation provides additional complementarity beyond exposing teachers to different training examples.

Unlike the benchmark-wise Teacher Envelope in Table~\ref{tab:main_results}, this analysis performs per-example aggregation over the cumulative teacher set.
Table~\ref{tab:residual_parallel_union} shows that residual GRPO produces a broader capability union. With four teachers, Residual GRPO achieves a Macro Avg.@8 of \(31.25\) and a Macro Pass@8 of \(59.29\), outperforming Parallel GRPO by \(2.71\) and \(6.61\) points, respectively. The difference is especially large on AIME: the four-teacher residual union improves Pass@8 from \(33.33\) to \(47.78\), an absolute gain of \(14.45\) points and a relative improvement of $43.4\%$. These results show that residual allocation is more effective than random partitioning at constructing complementary teachers.

The cumulative unions further clarify why. On AIME, the union of only two Residual-GRPO teachers already matches or exceeds the four-teacher Parallel-GRPO union on both Avg.@8 and Pass@8. On HMMT and AMC23, the residual union continues to improve as more teachers are added. This indicates that later teachers are not redundant; they expand the teacher union toward examples and solution modes that earlier teachers did not cover. Table~\ref{tab:residual_parallel_union} therefore supports the expansion side of our framework.

\paragraph{Distillation Objective: What Enables Effective Compression?}
We next compare the full framework with two variants.
\emph{Uniform Teacher-Union OPD} assigns equal weights to the available teachers and excludes the residual objective.
\emph{Quality-Weighted Teacher-Union OPD} uses per-example reliable
teachers and quality-aware weights, but removes
$\mathcal{L}_{\mathrm{res}}$.

Table~\ref{tab:opd_ablation_detailed} isolates the effect of the compression-stage design. We compare uniform Teacher-Union OPD, quality-weighted Teacher-Union OPD, and the full framework with Consensus-Residual Decomposition. We also vary the number of teachers used for distillation.

Quality-aware weighting provides a small but consistent aggregate benefit. Replacing uniform teacher aggregation with per-example quality weights improves the seven-benchmark math mean from \(15.81\) to \(15.99\). This suggests that teacher reliability is example-dependent and should not be treated uniformly.

The main improvement comes from preserving residual specialist signals. Adding the residual objective to the four-teacher quality-weighted OPD raises the mean from \(15.99\) to \(16.70\), a \(0.71\)-point gain. Compared with uniform Teacher-Union OPD, the full framework improves the mean by \(0.89\) points. This supports the role of Consensus-Residual Decomposition: standard teacher aggregation primarily transfers behavior shared by reliable teachers, while the residual objective preserves winner-specific preferences that would otherwise be weakened by averaging.

Teacher-set size also matters. Under the same full objective, using four teachers improves the mean from \(16.10\) to \(16.70\) compared with using two teachers. This shows that the additional teachers contribute useful information beyond the first two residual specialists. The gains are not uniform on every benchmark, but the best aggregate result is obtained only when both ingredients are present: a larger teacher union and residual transfer. 

Together, these results show that quality-aware aggregation and residual transfer play complementary roles: the former improves the reliability of the standard OPD signal by emphasizing competent teachers for each example, while the latter transfers specialist-specific preferences that are not captured by the shared distillation objective.
Thus, Table~\ref{tab:opd_ablation_detailed} supports the compression mechanism of our method.

\begin{table*}[t]
\centering
\caption{
Ablation of the multi-teacher distillation objective on Qwen3-1.7B.
All results are Avg.@8 (\%).
$\#T$ denotes the number of teachers used for distillation.
Mean is the unweighted average over the seven mathematical benchmarks,
and $\Delta_{\mathrm{full}}$ is the difference from the full
four-teacher framework.
}
\label{tab:opd_ablation_detailed}

\small
\setlength{\tabcolsep}{3.0pt}
\renewcommand{\arraystretch}{1.10}

\sisetup{
    detect-weight=true,
    detect-family=true,
    table-number-alignment=center
}

\begin{adjustbox}{max width=\textwidth}
\begin{tabular}{
    l
    c
    c
    c
    *{8}{S[table-format=2.2]}
    S[table-format=-1.2]
}
\toprule

\multirow{2}{*}{Configuration}
& \multirow{2}{*}{$\#T$}
& \multirow{2}{*}{Teacher Weighting}
& \multirow{2}{*}{$\mathcal{L}_{\mathrm{res}}$}
& \multicolumn{3}{c}{AIME}
& \multicolumn{3}{c}{HMMT}
& \multicolumn{1}{c}{\multirow{2}{*}{AMC23}}
& \multicolumn{1}{c}{\multirow{2}{*}{\makecell[c]{Math\\Mean}}}
& \multicolumn{1}{c}{\multirow{2}{*}{$\Delta_{\mathrm{full}}$}}
\\

\cmidrule(lr){5-7}
\cmidrule(lr){8-10}

& & &
& \multicolumn{1}{c}{2024}
& \multicolumn{1}{c}{2025}
& \multicolumn{1}{c}{2026}
& \multicolumn{1}{c}{Feb.\ 25}
& \multicolumn{1}{c}{Nov.\ 25}
& \multicolumn{1}{c}{Feb.\ 26}
& & &
\\

\midrule

Uniform Teacher-Union OPD
& 4
& Uniform
& No
& 10.83
& 12.92
& \bfseries 11.25
& 6.67
& \bfseries 8.75
& 8.71
& 51.56
& 15.81
& -0.89
\\

Quality-Weighted Teacher-Union OPD
& 4
& $\alpha_i^{(r)}$
& No
& 14.17
& 10.42
& 10.00
& \bfseries 7.50
& 7.92
& \bfseries 10.98
& 50.94
& 15.99
& -0.71
\\

\midrule

Full Framework
& 2
& $\alpha_i^{(r)}$
& Yes
& \bfseries 15.42
& 12.92
& 10.83
& \bfseries 7.50
& 7.92
& 10.61
& 47.50
& 16.10
& -0.60
\\

\textbf{Full Framework}
& 4
& $\alpha_i^{(r)}$
& Yes
& \bfseries 15.42
& \bfseries 15.00
& \bfseries 11.25
& 7.08
& \bfseries 8.75
& 7.20
& \bfseries 52.19
& \bfseries 16.70
& \bfseries 0.00
\\

\bottomrule
\end{tabular}
\end{adjustbox}
\end{table*}

\subsection{Model Scaling}

Table~\ref{tab:scaling_4b} evaluates whether the same behavior persists with a larger backbone. On Qwen3-4B, the final student reaches a Math Mean of \(21.97\), compared with \(20.55\) for the strongest Residual-GRPO teacher and \(15.73\) for the initialization policy. This corresponds to a \(6.9\%\) relative improvement over the strongest teacher and a \(39.7\%\) improvement over the initialization policy.

The improvement is consistent across benchmark families: the student improves over the strongest teacher by \(6.7\%\) on AIME, \(12.2\%\) on HMMT, and \(4.3\%\) on AMC23. These results suggest that the benefit of expanding and compressing a residual teacher union is not limited to the smaller Qwen3-1.7B setting. Even with a stronger initialization and stronger individual teachers, the teacher union still contains complementary capabilities that can be consolidated into a better single policy.

\begin{table}[t]
\centering
\caption{
Scaling results on mathematical reasoning with Qwen3-4B.
AIME and HMMT are averaged over their respective three benchmark.
}
\label{tab:scaling_4b}

\small
\setlength{\tabcolsep}{3.5pt}
\renewcommand{\arraystretch}{1.12}

\sisetup{
    detect-weight=true,
    detect-family=true,
    table-number-alignment=center
}

\begin{tabular}{l *{4}{S[table-format=2.2]}}   % 去掉 tabular* 和 \extracolsep
\toprule
Model
& \multicolumn{1}{c}{AIME}
& \multicolumn{1}{c}{HMMT}
& \multicolumn{1}{c}{AMC23}
& \multicolumn{1}{c}{\makecell[c]{Math Mean}} \\
\midrule
Initialization Policy ($\pi_0$)
& 13.20
& 6.00
& 52.50
& 15.73 \\
Best Residual Teacher
& 18.61
& 10.16
& 57.50
& 20.55 \\
\textbf{Ours}
& \bfseries 19.86
& \bfseries 11.40
& \bfseries 60.00
& \bfseries 21.97 \\
\bottomrule
\end{tabular}
\end{table}

\section{Conclusion}

We introduced an expand-then-compress framework for consolidating complementary capabilities from RL-trained reasoning policies.
Instead of treating a single RL policy as a globally complete teacher, we view each policy as a local specialist over the reasoning solution manifold.
Residual GRPO expands the capability union by directing successive teachers toward examples not yet covered by the accumulated teacher set.
Reliability-Gated Teacher-Union OPD then compresses this union on student-generated prefixes, while Consensus-Residual Decomposition preserves specialist-specific preferences that conventional teacher aggregation may suppress.

Across mathematical reasoning, code generation, and instruction following, the resulting Qwen3-1.7B student surpasses the strongest individual teacher by relative margins of $2.0\%$, $8.3\%$, and $6.9\%$, respectively, while retaining single-model inference.
Residual GRPO further improves the four-teacher capability union over static random partitioning by $2.71$ Macro Avg.@8 and $6.61$ Macro Pass@8 points, and the Qwen3-4B student maintains a $6.9\%$ relative advantage over its strongest teacher on mathematical reasoning.
Together, these results validate both stages of the framework: coverage-aware RL constructs genuinely complementary specialists, and residual-aware OPD converts their union into a student that outperforms every individual teacher on the primary aggregate metric of each task domain.
While our current study focuses on verifiable tasks and a fixed teacher set, the central principle is broader: \emph{build complementary teachers first, then distill the union rather
than the average.}

\bibliographystyle{plainnat}
\bibliography{aaai27_lss}

\end{document}